\documentclass[letterpaper]{article} 
\usepackage{aaai25}  
\usepackage{times}  
\usepackage{helvet}  
\usepackage{courier}  
\usepackage[hyphens]{url}  
\usepackage{graphicx} 
\usepackage{natbib}  
\usepackage{caption} 
\usepackage{algorithm}
\usepackage{algorithmic}

\usepackage{newfloat}
\usepackage{listings}
\DeclareCaptionStyle{ruled}{labelfont=normalfont,labelsep=colon,strut=off} 
\floatstyle{ruled}
\newfloat{listing}{tb}{lst}{}
\floatname{listing}{Listing}
\usepackage[absolute,overlay]{textpos}
\usepackage{multirow}
\usepackage{cite}
\usepackage{amssymb}

\title{DSCformer: A Dual-Branch Network Integrating Enhanced Dynamic Snake Convolution and SegFormer for Crack Segmentation}

\author{
    Kaiwei Yu\textsuperscript{1}\thanks{Email: 12332394@mail.sustech.edu.cn}, I-Ming Chen\textsuperscript{2}, Jing Wu\textsuperscript{1}\thanks{Corresponding author: wujing099@gmail.com}
}
\affiliations{
    \textsuperscript{1}Department of Mechanical and Energy Engineering, Southern University of Science and Technology\\
    \textsuperscript{2}School of Mechanical and Aerospace Engineering, Nanyang Technological University
}

\begin{document}

\maketitle

\begin{abstract}
    In construction quality monitoring, accurately detecting and segmenting cracks in concrete structures is paramount for safety and maintenance. Current convolutional neural networks (CNNs) have demonstrated strong performance in crack segmentation tasks, yet they often struggle with complex backgrounds and fail to capture fine-grained tubular structures fully. In contrast, Transformers excel at capturing global context but lack precision in detailed feature extraction. We introduce DSCformer, a novel hybrid model that integrates an enhanced Dynamic Snake Convolution (DSConv) with a Transformer architecture for crack segmentation to address these challenges. Our key contributions include the enhanced DSConv through a pyramid kernel for adaptive offset computation and a simultaneous bi-directional learnable offset iteration, significantly improving the model's performance to capture intricate crack patterns. Additionally, we propose a Weighted Convolutional Attention Module (WCAM), which refines channel attention, allowing for more precise and adaptive feature attention. We evaluate DSCformer on the Crack3238 and FIND datasets, achieving IoUs of 59.22\% and 87.24\%, respectively. The experimental results suggest that our DSCformer outperforms state-of-the-art methods across different datasets.
    
\end{abstract}

%
\section{Introduction}
Monitoring the quality of concrete structures is crucial in industrial settings. With the aging of concrete buildings, such as pavement, building exteriors, bridges, etc., cracks inevitably appear, resulting in significant safety hazards \cite{lee2024unified}. Cracks in road surfaces potentially expand into large potholes during a rainy night, and bridge cracks may lead to structural collapse. In recent years, such catastrophic accidents resulting in significant casualties have occurred multiple times in countries like China and the United States \cite{zou2018deepcrack}. Therefore, regular crack inspection is very important, in which crack segmentation and subsequent quantitative assessment provide valuable information for the health analysis of concrete structures. Numerous researchers focus on developing more accurate crack segmentation methods. 
\begin{figure}[ht]
    \centering
    \includegraphics[scale=0.2]{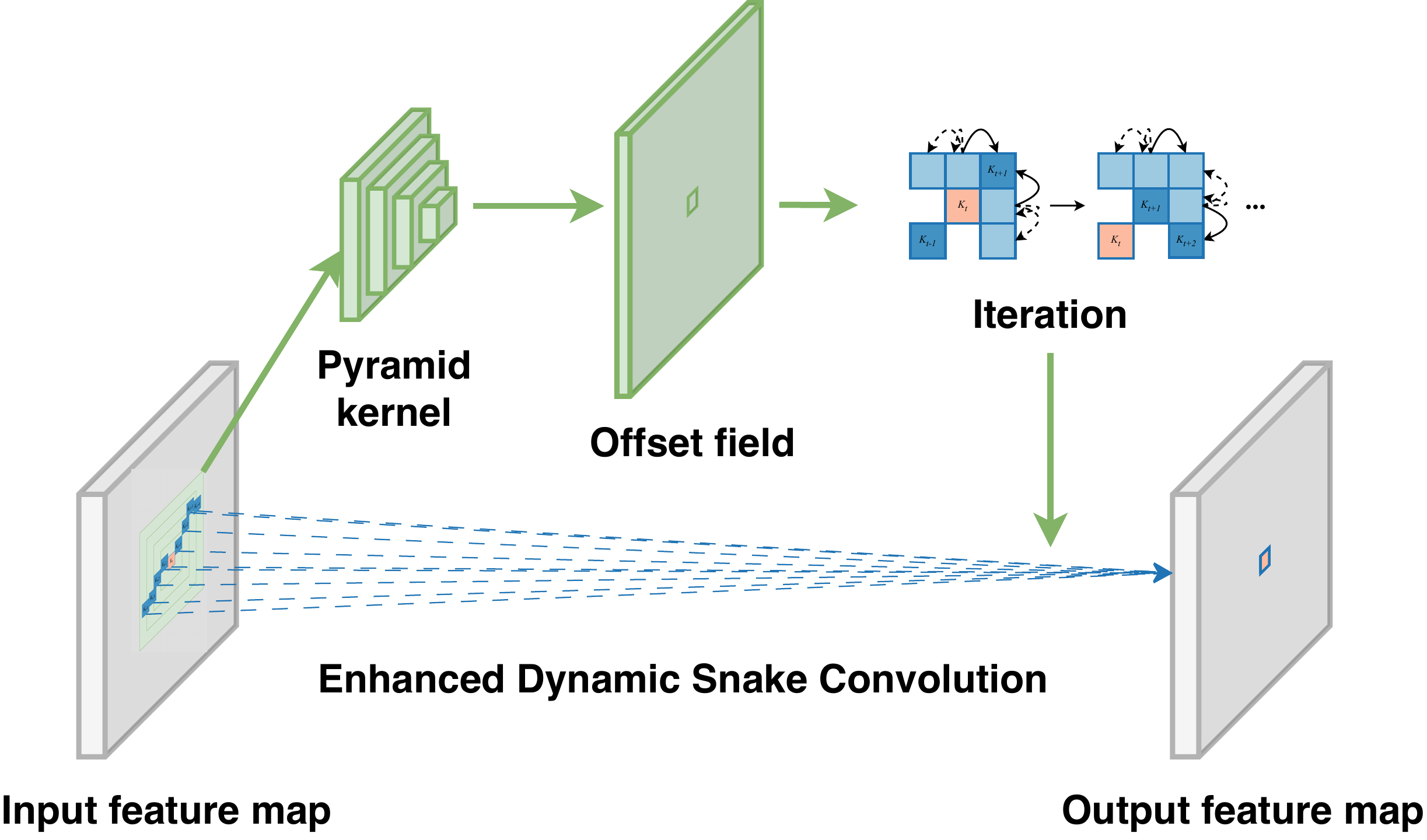}
    \caption{Overview of the Enhanced Dynamic Snake Convolution. A pyramid kernel is employed to obtain offsets on the input feature map. These offsets are iterated into chains through bi-direction, which is then convolved with the input feature map to gain the output feature map.}
\end{figure}
Traditional segmentation methods struggle with robustness due to background noise, such as lighting and graffiti. However, deep learning models have shown exceptional performance due to their accuracy, robustness and generalization.

Several studies \cite{liu2019deepcrack,cheng2021joint,zhang2022high} have employed Convolutional Neural Networks (CNNs) for crack segmentation. CNNs leverage the principle of local connectivity, where convolutional kernels focus on specific regions of an image. These kernels are particularly adept at detecting edges and textures, endowing the networks with strong translational invariance, which is advantageous for crack segmentation tasks. In addition, Convolutional attention mechanisms like Squeeze-and-Excitation (SE) \cite{hu2018squeeze} and the Convolutional Block Attention Module (CBAM) \cite{woo2018cbam} make the networks more focused on key information. Nevertheless, CNNs still struggle with complex backgrounds due to their limited receptive fields, leading to sub-optimal performance in such scenarios.

\begin{figure*}[t]
    \centering
    \includegraphics[scale=0.8]{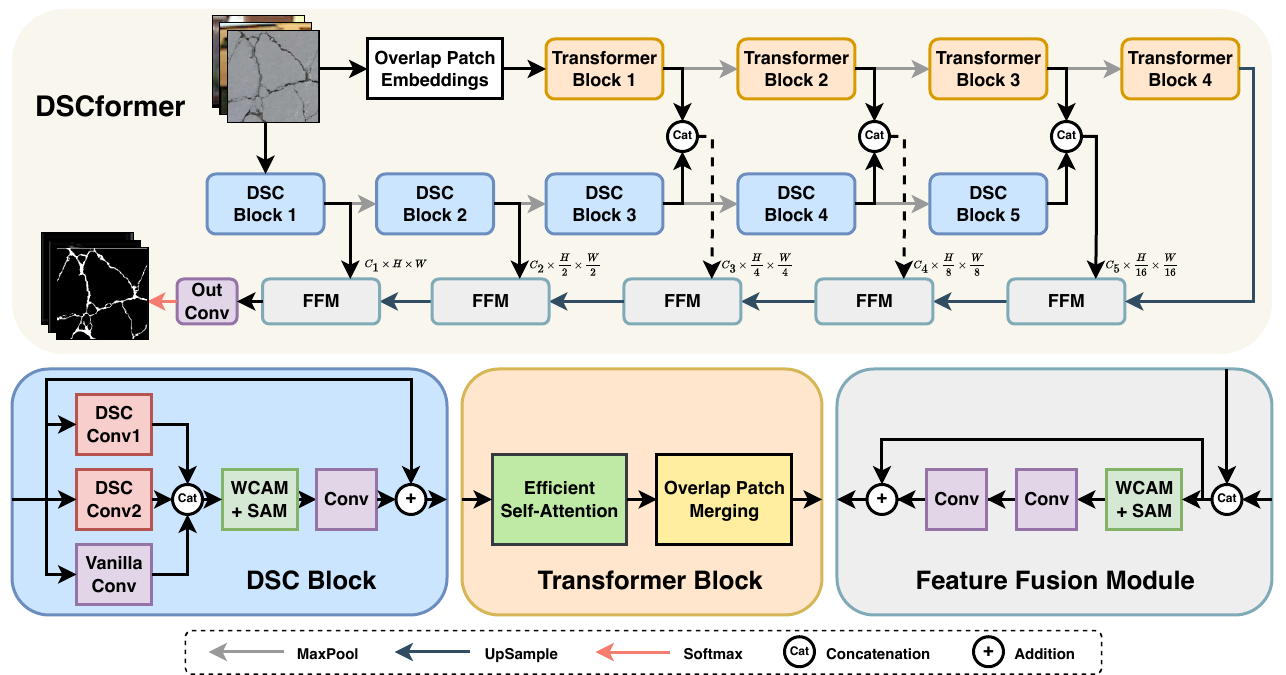}
    \caption{Illustration of the proposed DSCformer. We employ the enhanced dynamic snake convolution (DSConv) and Segformer as dual-branch encoder and use Weighted Convolutional Attention Module (WCAM) and Spatial Attention Module (SAM) to adjust the convolution focus.}
\end{figure*}

In contrast, Transformer networks offer a novel perspective by utilizing embeddings and self-attention mechanisms to operate on the entire input sequence simultaneously. This enables them to capture global information and model long-range dependencies effectively, which is beneficial for segmenting cracks in complex backgrounds \cite{shamsabadi2022vision,guo2023pavement}. However, Transformers often lack precision in identifying fine details.

The hybrid approach of combining CNNs and Transformers holds significant potential by leveraging both local detail extraction and global context modeling \cite{gao2021utnet,liu2021crackformer,wang2024dual}. However, the above methods often overlook the specific structure of crack segmentation when selecting convolutional and Transformer components, leading to inadequate fusion outcomes. Additionally, the integration of CNNs and Transformers can introduce complex training fluctuation, making it difficult to fine-tune the model to effectively handle both local details and global structures. Consequently, while these hybrid models have demonstrated performance improvements, they still requires further refinement to reach its potential.
In this context, Dynamic Snake Convolution (DSConv)\cite{qi2023dynamic} is very suitable for crack segmentation task. DSConv adaptively focuses on the fine and curved local features of tubular structures, enhancing geometric perception. Despite its effectiveness, the original DSConv has certain limitations in its flexibility and adaptability when dealing with complex structures, indicating room for improvement.

In this paper, we introduce DSCformer, a novel crack segmentation model that combines proposed enhanced DSConv with a Transformer-based architecture. Our approach introduces two key improvements to the original DSConv: a) a pyramid kernel for offset computation, consisting of convolutional kernels of varying sizes, and b) a method for learnable offset iteration that simultaneously operates in vertical and horizontal direction. Fig. 1 shows the process for enhanced DSConv. These enhancements allow the model to capture fine crack structures more accurately.
Furthermore, we have introduced the Weighted Convolutional Attention Module (WCAM), which refines channel attention by utilizing separate Multi-Layer Perceptrons for average and max pooling, aggregated through a weighted sum of learnable parameters. This modification facilitates more adaptive and precise channel attention.
The SAM remains unaltered, continuing to contribute spatial focus to the network. By integrating these modifications, we have developed the DSC block, which, when stacked in sequence, forms a DSConv encoder branch. This branch is aimed at yielding fine-grained and tubular topologically attentive feature maps. Additionally, we integrate a pre-trained SegFormer \cite{xie2021segformer} encoder branch to capture long-range dependencies. The features extracted by these two branches are fused and progressively upsampled to generate the final crack segmentation prediction map.
The principal contributions of this study are encapsulated as follows:
\begin{itemize}
    \item We propose an enhanced DSConv, incorporating a pyramid kernel for offset calculation and a simultaneous bi-direction learnable offset iteration, significantly improving ability to capture crack structure.
\end{itemize}
\begin{itemize}
    \item We propose WCAM, which refines the channel attention mechanism, improving its adaptability and precision.
\end{itemize}
\begin{itemize}
    \item We propose DSCformer, a hybrid model combining enhanced DSConv and SegFormer architectures, outperforming state-of-the-art crack segmentation methods across two public datasets.
\end{itemize}

\section{Related Works}
\subsection{Convolutional Networks for Crack Segmentation}
Early works utilized CNN architectures like Fully Convolutional Networks (FCNs) \cite{long2015fully}, U-Net \cite{ronneberger2015u}, and Feature Pyramid Networks (FPN) \cite{lin2017feature} become the basic framework for segmentation methods. DeepCrack uses the encoder-decoder structure based on SegNet \cite{badrinarayanan2017segnet} and is a classic work of crack segmentation. DcsNet introduces a detail branch that does not downsample layer by layer to construct detailed information. JTFN \cite{cheng2021joint} aims to model global topology and fine features simultaneously based on iterative feedback learning strategy. 

Among convolution-altered methods, Dilated Convolutions \cite{yu2017dilated} expanded the receptive field without losing resolution, allowing models to capture more context. Deformable Convolutions \cite{dai2017deformable} further advanced this by introducing learnable offsets to the convolutional kernels to adapt to the shape of features, which provided a stronger ability to perceive objects.
Cracks, akin to blood vessels and roads, exhibit fine tubular topological structures. Enhancing the perception of these structures can significantly improve segmentation accuracy. 
Recently, Dynamic Snake Convolution (DSConv) \cite{qi2023dynamic} was proposed, which limits the motion of deformable convolution sampling points, further enhances the geometric sensing ability of thin and curved tubular structures, and significantly improves the segmentation performance. However, DSConv has some limitations in crack perception due to its fixed offset kernel size and learnable offset iteration in a single-direction.
Specifically, using a fixed kernel size of 3 to calculate the offset may cause the snake chain of DSConv to iterate inaccurately away from the center point. The strategy of separate learnable offset iteration in one direction and fixed offset iteration in the another direction may lead to a lack of adaptive topology capability of the snake chain.

\subsection{Hybrid CNN-Transformer Models}
Vision Transformer (ViT) \cite{DBLP:conf/iclr/DosovitskiyB0WZ21} utilizes patch and position embedding and self-attention mechanisms to capture the global context of the entire image. After that, SETR \cite{zheng2021rethinking} performed intensive prediction tasks by using pure Transformer encoders. Swin Transformer \cite{liu2021swin} introduced a hierarchical approach with shifted windows for better computational efficiency and the ability to handle different image scales. Pyramid Vision Transformers (PVT) \cite{wang2021pyramid} combined the benefits of transformers with multi-scale feature extraction methods, making it well-suited for tasks requiring both local and global context. SegFormer \cite{xie2021segformer} introduces a hierarchical transformer encoder without position embedding and outputs multi-scale features. But pure Transformers have a limited ability to capture fine detail.

Recently, many studies have attempted to use a combination of CNN and Transformer for crack segmentation. Crackformer \cite{liu2021crackformer} applied a convolution layer to the top of the transformer encoder. UTNet \cite{gao2021utnet} introduced self-attention into the U-Net structure. UCTransNet \cite{wang2022uctransnet} proposed channel Transformer as U-Net skip connections. Both crackmer \cite{wang2024dual} and DTrC-Net \cite{xiang2023crack} use CNN and Transformer parallel encoder architecture. However, it still requires more advanced hybrid models to address the challenge of crack segmentation.

\subsection{Attention Mechanisms for Feature Fusion}
The Squeeze-and-Excitation (SE) network \cite{hu2018squeeze} recalibrates channel feature responses through average pooling, pioneering the concept of channel attention and enhancing the representational capacity of CNNs. Subsequently, the Convolutional Block Attention Module (CBAM) \cite{woo2018cbam} extends the attention mechanism by sequentially applying a Channel Attention Module (CAM) and a Spatial Attention Module (SAM), utilizing both average and max pooling information to enable the network to more accurately focus on relevant features and their locations. In the context of crack segmentation, crackmer introduces a Complementary Fusion Module to more effectively aggregate multi-scale features, leveraging complementary information from different feature levels to improve segmentation accuracy. Nevertheless, the fusion of max and average poolings information is not sufficient.

\section{DSCformer for Crack Segmentation}
Fig. 2 illustrates an overview of the DSCformer framework. To the best of our knowledge, current hybrid crack segmentation methods that combine CNN and Transformer architecture primarily focus on adding various feature fusion modules or designing more sophisticated decoders. However, we believe that the potential limitation of current hybrid models lies in feature extraction rather than feature fusion, which suffices for most tasks. As mentioned in the section on convolutional networks, conventional convolutions fail to fully capture the features of cracks or other fine-grained tubular structures. Although significant work, Dynamic Snake Convolution, attempted to enhance the perception of slender and tortuous local structures, issues with its offset calculation and iteration strategy may adversely affect the final performance. Building on and inspired by this work, we propose a dual-branch convolutional and transformer model, DSCformer. The convolutional branch is constructed by stacking DSC blocks, which include our proposed enhanced Dynamic Snake Convolution (DSConv) and Weighted Channel Attention (WCAM). The Transformer branch leverages the SegFormer encoder. Finally, crack segmentation masks are generated through progressive feature fusion and upsampling.
    \subsection{DSConv encoder branch: Using enhanced DSConv for Encoder Feature Extraction}
        The DSConv encoder branch is employed for local fine-grained structure feature extraction. We refine the computation strategy of DSConv, and propose a DSC block and WCAM. 
        \begin{figure}[ht]
            \centering
            \includegraphics[scale=0.7]{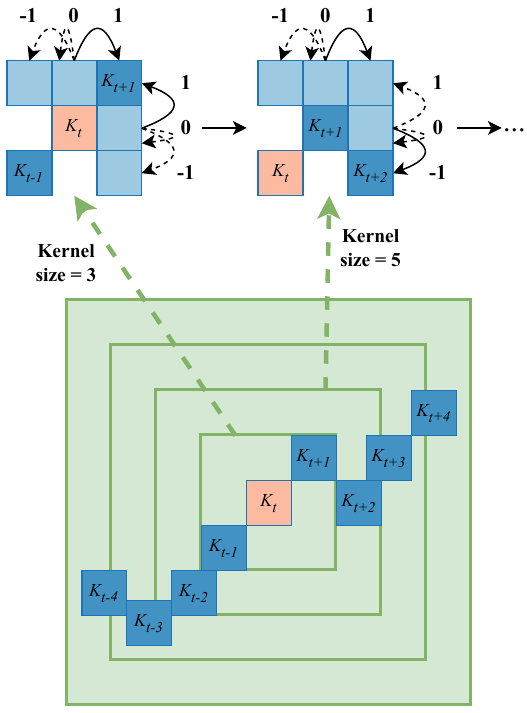}
            \caption{Offset iteration process. Iterating from the central point, each layer of the pyramid kernel represents its maximum receptive field.}
        \end{figure}
        The DSC block incorporates WCAM to enhance feature fusion between vanilla convolution and enhanced DSConv. By stacking DSC blocks and adding max-pooling layers between DSC blocks, we perform progressive downsampling to obtain multi-scale feature maps at resolutions of 1/1, 1/2, 1/4, 1/8, and 1/16.
        
        \subsubsection{Enhanced Dynamic Snake Convolution}
            In this section, we introduce the computation process of Dynamic Snake Convolution (DSConv) and present our improvements. Given a standard $3*3$ convolution kernel with a dilation of 1 as $K$, the kernel samples information from nine points: the central point $K_i=(x_i,y_i)$ and its surrounding eight points, which is expressed as:

            \begin{equation}
                K={(x-1,y-1),(x-1,y),...,(x+1,y+1)}
            \end{equation}
            
            To enable the convolution kernel to adapt to the tubular topological features around the current location, the goal is to allow the sampling points of the kernel to shift. Inspired by the concept of deformable convolutions, Qi et al. introduced deformation offsets $\Delta$ \cite{qi2023dynamic}. The offset convolution outputs offsets of 9 sampling points in the vertical and horizontal directions for a total of 18 channels. However, allowing the offsets to be completely learned freely can cause the receptive field to drift away from the target, which is unsuitable for capturing the characteristics of slender tubular structures. To address this, an iterative strategy is adopted to constrain the offsets, ensuring continuity in perception.

            Unlike previous method that utilizes a $3*3$ convolution kernel to compute the sampling point offsets, we propose a pyramid-structured convolution kernel, which contains kernel sizes of $3*3$, $5*5$, $7*7$, $9*9$. 
            \begin{figure}[ht]
                \centering
                \includegraphics[scale=0.2]{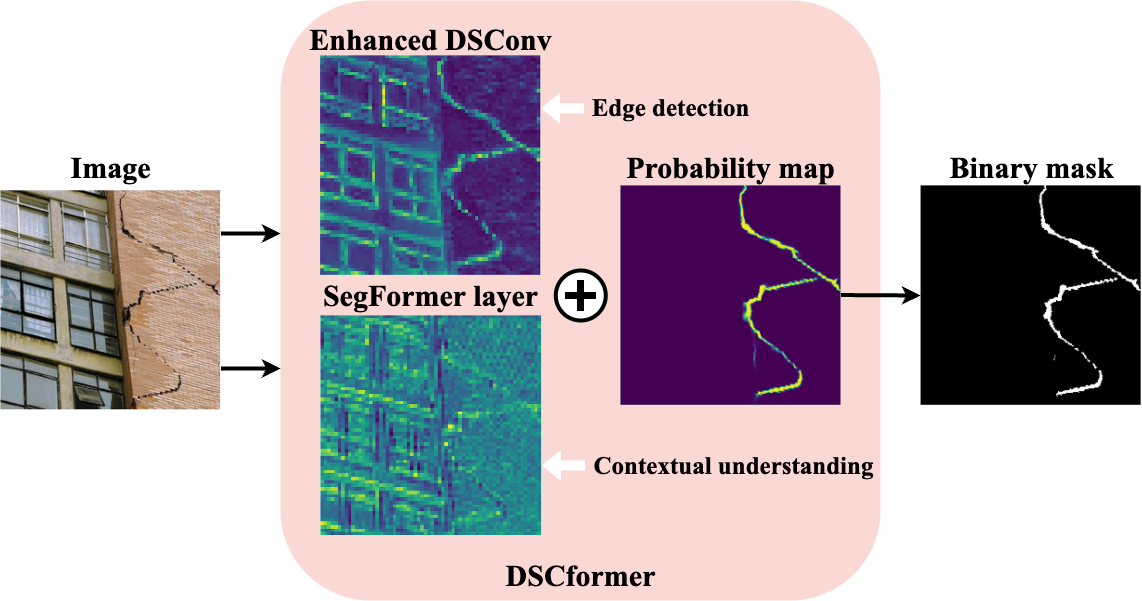}
                \caption{The difference between dual-branch encoder. The enhanced DSConv and Segformer layers are employed for edge detection and contextual understanding, respectively.}
            \end{figure}
            Specifically, the $3*3$ kernel in our design outputs offsets for points near the center of the dynamic snake convolution, while the larger sizes of kernel outputs offsets for points further away, up to the $9*9$ kernel. This approach gradually perceives information from the outer layers, matching the receptive field for offset acquisition with the actual sampling points, thereby avoiding limited perception issues.

            In the iterative offset strategy of DSConv, we consider a $9*9$ square, which is also its maximum receptive field. Each grid position in the kernel $K$ is represented as $K_{t\pm c}=(x_{t\pm c},y_{t\pm c})$, where $c={0,1,2,3,4}$. Unlike previous method, which fixes the offset as an integer in one direction while leaving the offset in the another direction unchanged, the enhanced DSConv do not fix the offset and iterate it in bi-direction simultaneously. So $t\pm c$ represents the iterative order of the convolution sampling points. Starting from the center point $K_t$, 

            \begin{equation}
                K_{t+c}=(x_{t+c},y_{t+c})=(x_t+\sum_{t}^{t+c} \Delta x,y_t+\sum_{t}^{t+c} \Delta y)
            \end{equation}

            \begin{equation}
                K_{t-c}=(x_{t-c},y_{t-c})=(x_t-\sum_{t-c}^{t} \Delta x,y_t-\sum_{t-c}^{t} \Delta y),
            \end{equation}

            After obtaining the coordinate of each sampling point after the offset iteration, the actual sampling point value is obtained by applying bilinear interpolation:
            
            \begin{equation}
                K={\sum_{K^{'}}} G(K^{'},K)\cdot K^{'}
            \end{equation}

            Where K represents the position of the coordinates, K' represents the grid point near K, and G(·, ·) is the bilinear interpolation kernel. Note that G is two dimensional. It is separated into two one dimensional kernels as:

            \begin{equation}
                G(K,K^{'})=g(K_x,K_{x}^{'})\cdot g(K_y,K_{y}^{'})
            \end{equation}
            
            As shown in Fig. 3, the enhanced DSConv improves the crack sensing ability of the snake chain away from the center point, and more flexibly captures the crack characteristics in the $9*9$ receiving field.

\begin{table*}[t]
\centering
\begin{tabular}{|l|l|c|c|c|c|c|c|}
\hline
                           \rule{0pt}{0.35cm}\textbf{Dataset} & \textbf{Methods} & \textbf{Param(M)} & \textbf{IoU (\%)$\uparrow$} & \textbf{Precision (\%) $\uparrow$} & \textbf{Recall (\%) $\uparrow$} & \textbf{F1 (\%) $\uparrow$} & \textbf{HD (mm) $\downarrow$} \\ \hline \hline
\multirow{7}{*}{Crack3238}  
\rule{0pt}{0.35cm}          & Unet              & $17.3$ & $53.43_{\pm 0.13}$          & $65.74_{\pm 0.57}$          & $67.67_{\pm 3.42}$          & $66.30_{\pm 0.15}$          & - \\ 
                            & DeepCrack         & $30.9$ & $54.22_{\pm 0.77}$          & $65.86_{\pm 1.29}$          & $72.06_{\pm 0.87}$          & $67.25_{\pm 0.89}$          & $56.28_{\pm 5.82}$ \\
                            & DSCNet & \hspace{2mm}$2.1$ & $53.99_{\pm 0.16}$          & $67.15_{\pm 1.69}$          & $70.45_{\pm 1.98}$          & $66.89_{\pm 0.19}$          & $41.15_{\pm 0.90}$ \\ 
                            & UCTransNet        & $66.5$ & $55.40_{\pm 0.31}$          & $67.61_{\pm 0.83}$          & $73.26_{\pm 0.38}$          & $68.52_{\pm 0.43}$          & $44.23_{\pm 1.82}$ \\ 
                            & DcsNet            & $15.2$ & $56.85_{\pm 0.09}$          & $69.83_{\pm 1.48}$          & $73.73_{\pm 1.37}$          & $70.05_{\pm 0.08}$          & $37.17_{\pm 0.73}$ \\ 
                            & DTrC-Net          & $41.8$ & $56.90_{\pm 0.33}$          & $70.17_{\pm 1.42}$          & $73.11_{\pm 1.51}$          & $70.06_{\pm 0.39}$          & - \\
                            & DSCformer(ours)   & $14.8$ & $\textbf{58.74}_{\pm 0.47}$ & $\textbf{72.38}_{\pm 2.22}$ & $\textbf{73.99}_{\pm 1.33}$ & $\textbf{71.74}_{\pm 0.48}$ & $\textbf{33.35}_{\pm 0.97}$ \\ \hline \hline
\multirow{7}{*}{FIND}  
\rule{0pt}{0.35cm}          & Unet              & $17.3$ & $85.57_{\pm 0.18}$          & $\textbf{92.97}_{\pm 0.61}$ & $91.65_{\pm 0.82}$          & $92.00_{\pm 0.11}$          & $13.21_{\pm 0.73}$ \\ 
                            & DeepCrack         & $30.9$ & $84.25_{\pm 0.16}$          & $91.40_{\pm 0.84}$          & $91.65_{\pm 0.88}$          & $91.21_{\pm 0.09}$          & $14.12_{\pm 0.96}$ \\
                            & DSCNet & \hspace{2mm}$2.1$ & $85.75_{\pm 0.07}$          & $92.35_{\pm 0.70}$          & $92.40_{\pm 0.67}$          & $92.12_{\pm 0.04}$          & $12.23_{\pm 0.20}$ \\ 
                            & UCTransNet        & $66.5$ & $86.05_{\pm 0.26}$          & $91.15_{\pm 0.29}$          & $93.98_{\pm 0.48}$          & $92.30_{\pm 0.16}$          & $11.90_{\pm 0.61}$ \\ 
                            & DcsNet            & $15.2$ & $85.02_{\pm 0.09}$          & $91.97_{\pm 1.00}$          & $91.98_{\pm 1.09}$          & $91.64_{\pm 0.06}$          & $14.67_{\pm 0.86}$ \\ 
                            & DTrC-Net          & $41.8$ & $74.28_{\pm 0.07}$          & $83.50_{\pm 0.54}$          & $84.94_{\pm 1.59}$          & $84.21_{\pm 0.02}$          & $14.76_{\pm 0.09}$ \\
                            & DSCformer(ours)   & $14.8$ & $\textbf{87.31}_{\pm 0.12}$ & $92.52_{\pm 0.35}$ & $\textbf{94.01}_{\pm 0.45}$ & $\textbf{93.04}_{\pm 0.09}$ & $\textbf{11.14}_{\pm 0.23}$ \\ \hline
\end{tabular}
\caption{Performance Comparison on Crack3238 and FIND Datasets with three test results.}

\end{table*}

        \subsubsection{Weighted Channel Attention Module}
            In order to better focus on more relevant channel information, we propose WCAM, which is based on CAM, and makes better utilization of average pooling and maximum pooling information by introducing a few parameters. Reduces the deviation of attention due to accidental maximum values.

            Mathematically, similarly, spatial information is first aggregated using max pooling and average pooling operations to obtain two distinct spatial identifiers: $F_{avg}^c, F_{max}^c$, each containing channel information. These two identifiers are then passed through independent MLPs, denoted as $M_{avg}(\cdot)=W_{avg}^{1}(W_{avg}^{0}(\cdot))$ and $M_{max}(\cdot)=W_{max}^{1}(W_{max}^{0}(\cdot))$ which, compared to a shared MLP, introduce only a small number of additional parameters with minimal impact on computational complexity. After applying the sigmoid activation function, the two output feature vectors are weighted, with the weights being learnable parameters. In summary, WCAM can be computed as follows:
            \begin{equation}
                M_c(F)=\sigma(w_{avg}M_{avg}(F_{avg}^c)+w_{max}M_{max}(F_{max}^c)) 
            \end{equation}
                
        
            Here, $\sigma$ denotes the sigmoid function, $F$ contains $F_{avg}^c$ and $F_{max}^c$, \(W_{avg}^{0}, W_{max}^{0} \in \mathbb{R}^{C/r \times C} \), \( W_{avg}^{1}, W_{max}^{1} \in \mathbb{R}^{C \times C/r} \), and $w_{avg}, w_{max} \in \mathbb{R}^{C}$. The ReLU activation function is followed by $W_{avg}^{0}, W_{max}^{0}$.

        \subsubsection{DSC block}
        In the DSConv branch, we propose the DSC block to extract features at different scales. Initially, the feature map is fed into three parallel convolutions: two enhanced DSConvs for extracting fine structural features especially and one standard convolution for capturing local features. The outputs of the three convolutions are concatenated along the channel dimension and then processed through WCAM and SAM to adjust the feature map values, focusing on important channel information and spatial positions. This is followed by a standard convolution layer for feature fusion. Finally, a residual connection is introduced to mitigate the issue of gradient vanishing \cite{he2016deep}.

    \subsection{SegFormer encoder branch and decoder}
        To address the challenge of capturing global information, which convolutional models often struggle with, we integrate a Transformer branch into our model. Specifically, we employ the encoder from SegFormer-b0, known for its efficient self-attention mechanism and Non-positional encodings, which is particularly advantageous for semantic segmentation tasks. This encoder, containing only three million parameters, offers a computationally lightweight solution while still providing robust feature extraction capabilities. Moreover, we utilize the pre-trained SegFormer-b0 model fine-tuned on the ADE20k dataset \cite{zhou2017scene} to leverage prior knowledge about various background contexts, thereby aiding the DSConv branch in more accurately segmenting cracks. The SegFormer encoder outputs multi-scale feature maps at 1/4, 1/8, 1/16, and 1/32 of the original resolution.

        For the decoder, we fuse the feature maps obtained from the DSConv branch and the SegFormer encoder branch, along with upsampled features from the lower-resolution maps. These fused features are then processed through WCAM and SAM in the same way to refine the feature map values, focusing on the most relevant channel information and spatial locations. The refined features are subsequently passed through a series of two convolutional layers to smooth the decoded image, with residual connections added to mitigate the vanishing gradient problem. 
        As shown in Fig. 4, enhanced DSConv is more suitable for extracting edges and fine structures, while the SegFormer layer is better at global modeling. The final crack mask is obtained by ultimately fusing feature maps that contain different types of information.

\section{Experiments}
\begin{figure*}[t]
    \centering
    \includegraphics[scale=1]{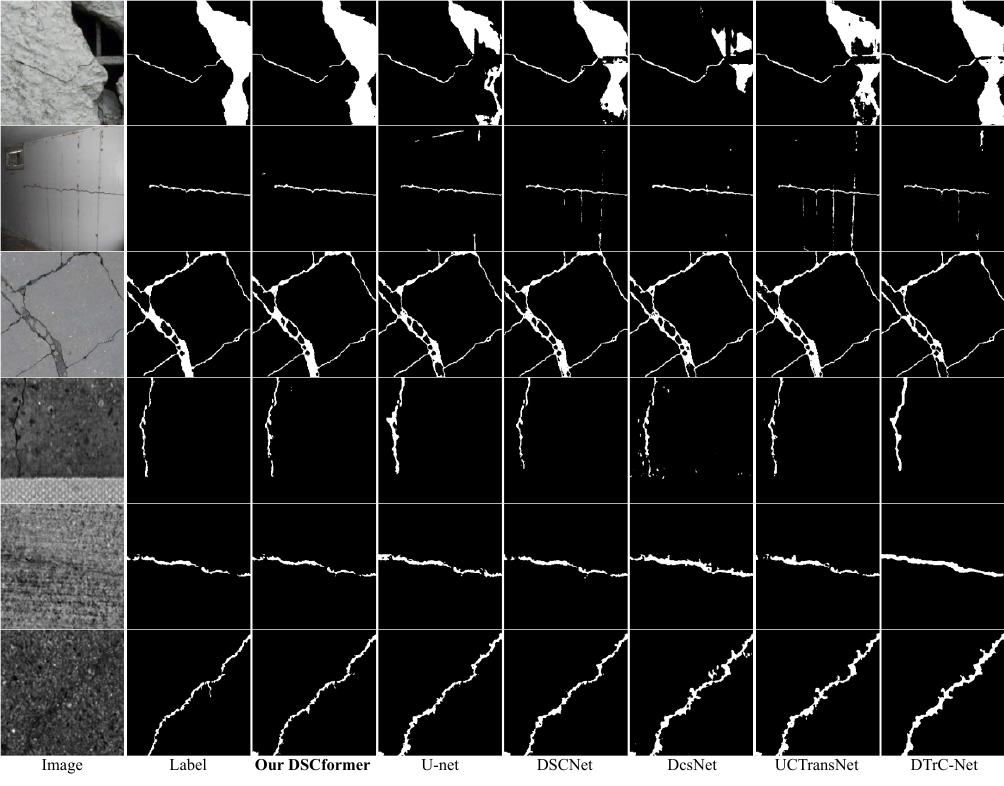}
    \caption{The qualitative comparison on the Crack3238 and FIND datasets.}
\end{figure*}
    \subsection{Datasets}
        We use Crack3238 dataset \cite{xiang2023crack}, Fused Image dataset for convolutional neural Network-based crack Detection(FIND) dataset \cite{Zhou2022FIND} to evaluate our method. Crack3238 dataset contains a total of 3,238 images, which consist of RGB images of cracks from various sources, featuring complex backgrounds, diverse crack patterns, and defective surfaces, significantly increasing the complexity of crack detection scenarios. The FIND dataset comprises 2,500 images, where fused image data is obtained by combining depth information with grayscale images. Both datasets encompass data from different sources and highly complex scenes, making them extremely challenging for crack segmentation tasks. We split the datasets, using \( 80\% \) for training and \( 20\% \) for testing.
    
    \subsection{Implementation Details}
        \subsubsection{Training details.}
        Our proposed model is implemented using the PyTorch framework. The training is performed on a single NVIDIA RTX 3090Ti GPU. To mitigate overfitting, we applied a variety of data augmentation techniques, including horizontal and vertical flips, random rotations, and affine transformations. The model incorporates pre-trained weights from SegFormer, as the SegFormer encoder branch leverages prior knowledge about the background surfaces (such as walls or road surfaces) where cracks typically occur. In contrast, DSConv encoder branch and decoder do not utilize pre-trained weights. We employ the Adam optimizer with a learning rate of 0.0001 and a weight decay of $10^{-4}$. We also combine cross entropy loss and dice loss as loss functions to train our network. The model was trained for 100 epochs with a batch size of 8.

        \subsubsection{Performance metrics.}
        To evaluate the performance of crack segmentation at the pixel level, we utilized several commonly used metrics: Intersection over Union (IoU), Recall, Precision, and F1-score. Their mathematical formulations are as follows: IoU is given by $ IoU = \frac{TP}{TP + FP + FN} $, Recall is given by $ Recall = \frac{TP}{TP + FN} $, Precision is given by $ Precision = \frac{TP}{TP + FP} $, and F1-score is given by $ F1=\frac{2 \times Precision \times Recall}{Precision + Recall} $. Additionally, we employed the Hausdorff Distance, calculated as $ d_H(A, B) = \max \left( \max_{a \in A} \min_{b \in B} |a - b|, \max_{b \in B} \min_{a \in A} |a - b| \right) $, to measure the maximum distance between the segmentation mask and the ground truth. All metrics for each image are calculated and averaged.

    \subsection{Comparison with State-of-the-art Methods}
        To demonstrate the overall segmentation quality of the proposed DSCformer, we compare it with other state-of-the-art methods. We evaluate three convolution-based models: U-Net, DeepCrack and DcsNet, hybrid models combining convolution and Transformer: UCTransNet and DTrCNet, and specifically, DSCNet, based on deformable convolution. For a fair comparison, we use their original code and published settings. To make the results more convincing, we test each model three times. The experimental results are presented in Table 1, where the metrics performance is shown in the form of mean plus standard deviation, and the best results are highlighted in bold. As shown in Table 1, our method outperforms all others in segmentation performance. On the Crack3238 dataset, DSCformer achieves an IoU of 58.74\%, Precision of 72.38\%, Recall of 73.99\%, and F1-score of 71.74\%. Similarly, on the FIND dataset, DSCformer obtains outstanding performance with an IoU of 87.31\% and an F1-score of 93.04\%, reaching the highest level. Although Precision and Recall are mutually exclusive metrics, our method balances them well, Precision achieves the second-best performance. 

        Additionally, we present a visual comparison of the segmentation results produced by the evaluated models in Fig. 5. This figure serves as a crucial point of reference, illustrating the performance differences among the models when applied to crack segmentation. Our method, DSCformer, stands out by markedly reducing the influence of potential noise. This robustness is particularly distinct in large defect areas, where DSCformer consistently outperforms other models by accurately capturing the extent and shape of these regions.
        
        Moreover, DSCformer specializes in delineating the crack edges, which is a critical aspect of crack segmentation. While many models tend to overestimate the width of cracks to achieve higher precision, DSCformer maintains precision by producing crack widths that closely match the ground truth. This results in a finer and more accurate segmentation, particularly at the boundaries, where even slight deviations can lead to significant differences in the final segmentation output. This level of detail highlights the model's strong generalization capabilities across varying defect types and sizes.
        Based on the comprehensive quantitative and visual analysis, it is clear that DSCformer outperforms the state-of-the-art models. Its ability to handle noise, precisely segment large defect areas, and accurately delineate crack edges suggests that DSCformer is a highly effective method for this downstream segmentation task.

\subsection{Ablation Studies}
We conducted ablation experiments on the Crack3238 dataset to assess the impact of the proposed improvements on our model's performance. As summarized in Table 2, the dual-branch architecture of DSCformer exhibits a significant advantage over the single-branch models, DSCNet and SegFormer. Specifically, DSCformer achieved an IoU improvement of 4.75\% over DSCNet and 4.07\% over SegFormer, highlighting the performance of integrating both branches for crack segmentation.
The enhanced DSConv also demonstrates substantial benefits. 
\begin{table}[h]
\begin{tabular}{|l|l|l|c|c|}
\hline
\rule{0pt}{0.35cm}\textbf{Methods}            & \textbf{Conv}   & \textbf{CA} & \textbf{IoU} & \textbf{F1} \\ \hline \hline
\rule{0pt}{0.35cm}DSCNet                     & DSConv          & w/o           &    53.99     &       66.89        \\ \hline
\rule{0pt}{0.35cm}SegFormer                  & w/o               & w/o           &    54.67          &    68.65           \\ \hline
\multirow{5}{*}{DSCformer} \rule{0pt}{0.35cm}& Vanilla Conv    & WCAM       &     57.46         &    70.51           \\ \cline{2-5} 
                           \rule{0pt}{0.35cm}& DSConv          & WCAM       &     57.74         &    70.82           \\ \cline{2-5} 
                           \rule{0pt}{0.35cm}& Enh. DSConv & w/o           &     58.16         &    71.09           \\ \cline{2-5} 
                           \rule{0pt}{0.35cm}& Enh. DSConv & CAM          &     58.28        &     71.22         \\ \cline{2-5} 
                           \rule{0pt}{0.35cm}& Enh. DSConv & WCAM       &     58.74         &    71.74          \\ \hline
\end{tabular}
\caption{Ablation experiments on Crack3238 datasets. `Enh. DSConv' denotes Enhanced Dynamic Snake Convolution. `CA' denotes the Convolutional Attention.}
\label{tab:ablation}
\end{table}
By incorporating the enhanced DSConv into the model, we observed a notable 1.00\% increase in IoU compared to the original DSConv, indicating that the improvements to the DSConv effectively contribute to more accurate crack boundary detection and segmentation.
Additionally, the introduction of WCAM provided further gains in segmentation performance. When WCAM was applied within DSCformer, it outperformed the standard CAM, resulting in a 0.46\% increase in IoU. This improvement reflects the value of our modifications to the attention mechanism.
Overall, these results emphasize the significance of the hybrid model architecture and the specific enhancements to both the dynamic snake convolution and channel attention mechanisms. Each modification contributes to the overall effectiveness of DSCformer, leading to superior crack segmentation performance.

\section{Conclusion}
    In this work, we presented DSCformer, a novel hybrid model that integrates enhanced Dynamic Snake Convolution (DSConv) and Transformer-based architectures for crack segmentation. By addressing the limitations of traditional CNNs and Transformers, DSCformer effectively captures both fine-grained local details and global context, making it particularly suitable for the challenging task of crack segmentation. Our enhanced DSConv, utilizing a pyramid kernel and bi-directional learnable offset iteration, significantly improves the model's ability to detect the features of crack structures. Additionally, the introduction of the Weighted Convolutional Attention Module (WCAM) further refines channel attention, enhancing the adaptability of feature fusion. Experimental results demonstrate that DSCformer outperforms state-of-the-art methods on multiple benchmark datasets. Moving forward, we believe that the integration of such hybrid architectures and enhanced DSConv will pave the way for more robust and accurate crack segmentation methods, with potential applications extending beyond construction quality monitoring to other domains requiring fine topological structure analysis.

\end{document}